\pdfoutput=1

\documentclass[11pt]{article}

\usepackage{acl}

\usepackage{times}
\usepackage{latexsym}

\usepackage{float}

\usepackage{microtype}

\usepackage[linesnumbered,ruled,vlined]{algorithm2e}
\SetKwInput{KwInput}{Input} 
\SetKwInOut{KwOutput}{Output} 

\definecolor{hotpink}{HTML}{EF7C8E}
\definecolor{tiffanyblue}{HTML}{A0E7E5}
\definecolor{mint}{HTML}{B4F8C8}
\definecolor{paleyellow}{HTML}{FBE7C6}
\definecolor{rosewater}{HTML}{D8A7B1}
\definecolor{cream}{HTML}{FAE8E0}

\usepackage[utf8]{inputenc} 
\usepackage[T1]{fontenc}    
\usepackage[russian,english]{babel}
\usepackage{hyperref}       
\usepackage{url}            
\usepackage{amsfonts}       
\usepackage{nicefrac}       
\usepackage{microtype}      
\usepackage{xcolor}         

\usepackage{booktabs}
\usepackage{amsmath}
\usepackage{multirow}
\usepackage{graphicx}
\usepackage{enumitem}
\usepackage{subcaption}
\usepackage{caption}
\usepackage{rotating}
\usepackage[normalem]{ulem}
\usepackage{amssymb}
\usepackage{colortbl}
\usepackage{linguex}

\usepackage{siunitx} 
\usepackage{multirow, makecell}
\usepackage{pifont}
\usepackage{siunitx} 
\usepackage{multirow, makecell}
\usepackage{tabularx}
\usepackage[all]{nowidow}
\definecolor{darkgreen}{rgb}{0.0, 0.42, 0.24}
\definecolor{green}{RGB}{112, 173,71}
\definecolor{blue}{RGB}{68, 114,196}
\definecolor{orange}{RGB}{237, 125,49}
\definecolor{red}{RGB}{202, 54,49}
\definecolor{yellow}{RGB}{222,194, 142}

\newcommand{\knnmt}{\textsc{kNN-MT}\xspace}

\usepackage{tcolorbox}
\usepackage{adjustbox}
\definecolor{azure}{rgb}{0.0, 0.5, 1.0}

\newcommand{\method}[1]{\textsc{DiPMT}}
\newcommand{\hint}[1]{\textsc{Hint}}
\definecolor{Gray}{gray}{0.9}
\title{Dictionary-based Phrase-level Prompting of Large Language Models\\ for Machine Translation}

\author{Marjan Ghazvininejad\qquad \qquad Hila Gonen \qquad \qquad Luke Zettlemoyer \\ \\ Meta AI}
\begin{document}
\maketitle
\begin{abstract}

Large language models (LLMs) demonstrate remarkable machine translation (MT) abilities via prompting, even though they were not explicitly trained for this task. However, even given the incredible quantities of data they are trained on, LLMs can struggle to translate inputs with rare words, which are common in low resource or domain transfer scenarios. 
We show that LLM prompting can provide an effective solution for rare words as well, by using prior knowledge from bilingual dictionaries to provide control hints in the prompts. 
We propose a novel method, \method{}, that provides a set of possible translations for a subset of the input words, thereby enabling fine-grained phrase-level prompted control of the LLM. 
Extensive experiments show that \method{} outperforms the baseline both in low-resource MT, as well as for out-of-domain MT. We further provide a qualitative analysis of the benefits and limitations of this approach, including the overall level of controllability that is achieved. 

\end{abstract}

\section{Introduction}

Large language models (LLMs) can be prompted to perform very high quality machine translation (MT), even though they were not explicitly trained for this task~\cite{brown2020language, lin2021few, zhang2022opt,scao2022bloom}. To do this, we simply ask them to complete a string \textit{Translate the following sentence to English} or \textit{French} etc. following by an input sentence. However, despite the fact that they are trained on massive corpora, these models can struggle to correctly translate rare words, which are common in low resource or domain transfer scenarios. In this paper, we show that the controlability that comes with prompting LLMs extends to the word level, allowing us to not only specify the overall task but also provide hints or options about individual choices the model should consider when performing the task. 

More specifically, we assume access to prior knowledge from bilingual dictionaries for machine translation, and show how to prompt LLMs with hints that specify a set of possible translation options for specific input words. As seen in Figure~\ref{fig:ex}, this involves simply appending a string such as \textit{in this context, the word ``membatasi'' means ``limiting,'' ``restrict,'' ``limit.''} to the end of the typical MT prompt, for every word that we have hints for in the lexicon. This approach is inspired by supervised machine translation models that have successfully used dictionaries to improve translation \cite{zhang2016bridging,arthur2016incorporating,zhong2020look}, but crucially we show how to incorporate the knowledge in a zero-shot fashion with no model training. It provides a simple method to not only specify the task in the prompt, but also provide background knowledge that might be useful for helping the model complete the task, without enforcing any hard constraints on how the model uses this knowledge.



Extensive experiments show that this approach, which we call \method{} (\textbf{Di}ctionary-based \textbf{P}rompting for \textbf{M}achine \textbf{T}ranslation), we are able to achieve significant gains for low-resource translation from and into English across multiple languages and language models. Additionally, we experiment also with out-of-domain translation, where we automatically extract dictionaries from the training data \cite{shi2021bilingual}, and show improvements of up to 13 Bleu points in this setting. As a next step into understanding the abilities of \method{} better, we analyze and explore the behaviour and the limits of the model, and measure the extent to which \method{} is controlling the final translation in practice. 

Our contributions in this work are the following: (a) we propose \method{}, a novel method for incorporating bilingual dictionary information in prompting-based machine translation (Section~\ref{sec:method}); (b) we show empirically that \method{} improves translation significantly both for low-resource and out-of-domain translation (Section~\ref{sec:improving}); (c) we analyze \method{} and provide insights about the benefits and limitations of the method (Section~\ref{sec:analysis}); (d) we explore  the extent to which \method{} controls the final translation in practice (Section~\ref{sec:control}).

\section{Machine Translation via Prompting with Dictionary}
\label{sec:method}

Prompting language models for translation assumes that the model is pretrained on enough training data in both the source and the target languages. However, that is often not the case, especially for low-resource languages and more so when the model is trained primarily on English data. Another challenge that affects translation quality is that of out-of-domain data. When asked to translate data that is out-of-domain with respect to the pretraining data, the model struggles to output high quality translations~\cite{koehn2017six}.


To alleviate these challenges, we present a method for incorporating dictionary knowledge into prompting-based MT. Dictionaries are relatively easy to obtain, even for low resource languages, making them appealing candidates for external source of translation \cite{zhang2016bridging, arthur2016incorporating, hamalainen2019template, zhong2020look}. Our method is simple, easy to use, and applies to any language pair where a dictionary exists. 


Our method, \method{} uses prompting-based translation and incorporates dictionary information into the prompt directly. Given a source sentence, we look for dictionary entries for the words included in it, and add them to the prompt directly: after asking for the translation of the full sentence, we add an additional part to the prompt that lists the possible translations for specific words. The prompt for each sentence has three parts as demonstrated in Figure~\ref{fig:ex}:

(1) the source sentence: ``\textit{Translate the following sentence to English: <source-sentence>}''; (2) the dictionary translations: ``\textit{In this context, the word X means A; the word Y means B,C,D.}''; (3) Asking for the translation to the target language: ``\textit{The full translation to English is:}''

To make the model familiar with the specific framework we choose, we add $k$ demonstrations before prompting for the current instance. These demonstrations include the translation of the source sentence in their third part. 

The dictionaries we use are word-level dictionaries. Extending \method{} to incorporate phrase-level translations should be trivial, and we leave this exploration to future work.

\begin{figure}[h!]
    \centering
    \begin{tcolorbox}[boxsep=0pt,left=0pt,colframe=white]
    \begin{tabular}{p{\linewidth}}
    
    \textcolor{violet} {Translate the following sentence to English: \textit{Pada dasarnya, hal tersebut terbagi ke dalam dua kategori: Anda bekerja sambil mengadakan perjalanan atau mencoba mencoba atau membatasi pengeluaran Anda. Artikel ini berfokus pada hal yang terakhir.}} \\ 
    \textcolor{azure} {In this context, the word "sambil" means "while"; the word "membatasi" means "limiting", "restrict", "limit".} \\ 
    \textcolor{green} {The full translation to English is: \textit{Basically, they fall into two categories: Either work while you travel or try and limit your expenses. This article is focused on the latter.}}\\ \\
    
    \textcolor{violet} {Translate the following sentence to English: \textit{Ia melakukan pembuatan bel pintu dengan teknologi WiFi, katanya.}} \\
    \textcolor{azure} {In this context, the word "pembuatan" means "creation"; the word "bel" means "buzzer", "bell"; the word "pintu" means "door", "doors".} \\
    \textcolor{green} {The full translation to English is:}
    \end{tabular}
    \end{tcolorbox}    
    \caption{An example of 1-shot prompt for Indonesian to English Translation. Each example consists of 3 parts: in violet -- the source sentence (along with the translation instruction and the target language), in blue -- the dictionary-based word-level translations, in green -- the translation to the target language. For the current instance to be translated, the third part is partial, and the model is expected to generate the translation into the target language. For the baselines, we use the same fomulation, but omit the second (blue) part of dictionary-based translations.}  
    \label{fig:ex}
\end{figure}

\method{} is simple and straight-forward to use, and also effective in improving translation quality, as we show in Section~\ref{sec:improving}. 

\section{Experimental Setup}

\subsection{Language Models}
We experiment with two large scale language models: OPT~\cite{zhang2022opt} which is mostly trained on English data, and Bloom~\cite{scao2022bloom}, a multilingual language model. 
\paragraph{OPT} We use the publicly available checkpoint of the $\text{OPT-175B}$, a decoder-only causal language model, with 96 layers and a hidden dimension of 12,288. 
The training corpora contains predominantly English text with a small amount of non-English data that has skipped the filtering process. Yet, the model exhibits good performance on few-shot translation, especially when translating into English (see Table~\ref{tab:indomain_results}, second column, for reference).


\paragraph{‌Bloom} 
To assess the effectiveness of our proposed method on multilingual language models, we also use the publicly available $\text{Bloom-176B}$ checkpoint. Bloom is trained on the ROOTS corpus~\cite{laurenccon2022bigscience} consisting of 498 Hugging Face datasets~\cite{lhoest2021datasets} involving 46 natural languages and 13 programming languages. Bloom is also a decoder-only transformer language model, with 70 layers and a hidden dimension of 14,336.

\subsection{Datasets and Evaluation Metrics}\label{sec:langs}

For in-domain evaluation, we use \text{Flores-101}~\cite{goyal2022flores}, which contains 3,001 sentences taken from English Wikipedia on various topics and domains. These sentences have been professionally translated into 101 different languages. As we aim to focus on low-resource languages (with respect to the model), we select 10 languages on which OPT performs moderately well\footnote{Based on a random selection of 60 examples from the development set} ($10-30$ BLEU points) in a four-shot machine translation setting, from and into English. These languages are listed in Table~\ref{tab:languages}. We refrain from using languages on which OPT performs poorly ($< 10$ BLEU points) since we assume that the performance for those ones is too low to expect reasonable translations even when incorporating external information.


\begin{table}[!t]
    \centering
    \resizebox{\columnwidth}{!}{
    \begin{tabular}{llll}
    \rowcolor{gray!10}
    \textbf{ISO 639-3} & \textbf{Language}  & \textbf{Family}  & \textbf{Subgrouping} \\
    cat & \textbf{Catalan} & Indo-European & Romance \\
    hrv & \textbf{Croatian} & Indo-European &  Balto-Slavic  \\
    dan & \textbf{Danish} & Indo-European & Germanic \\
    nld & \textbf{Dutch} & Indo-European & Germanic \\
    tgl & \textbf{Filipino} & Austronesian  & Austronesian \\
    ind & \textbf{Indonesian} & Austronesian  & Austronesian\\
    ita & \textbf{Italian} & Indo-European & Romance \\ 
    msa & \textbf{Malay} & Austronesian  & Austronesian\\ 
    nob & \textbf{Norwegian} & Indo-European  & Germanic \\ 
    slk & \textbf{Slovak} & Indo-European  &  Balto-Slavic\\

    \end{tabular}}
    \caption{The low-resource languages that are chosen from \text{Flores-101}~\cite{goyal2022flores} along with their language family and subgrouping attributes.}
    \label{tab:languages}
\end{table}

For out-of-domain evaluation we use $De-En$  data from \citet{aharoni-goldberg-2020-unsupervised}, covering the following domains: Medical, Law, IT, and Koran. The dataset statistics are presented in table~\ref{tab:out_of_domain}. All sentences longer than 250 tokens and sentence pairs with a source/target length ratio of more than 1.5 are removed from the training sets. 

We evaluate the detokenized length generated by the model using sacreBLEU \cite{post-2018-call}.\footnote{\url{https://github.com/mjpost/sacrebleu}}

\begin{table}[h]
\centering
\begin{tabular}{lrrr}
 \toprule
 \textbf{Dataset}  & \textbf{Train}  & \textbf{Dev} & \textbf{Test} \\
\midrule
  Medical & $248$K & 2000& 2000\\
  Law & $467$K & 2000& 2000\\
  IT & $223$K & 2000& 2000\\
  Koran & $17$K & 2000& 2000\\
  \bottomrule
 \end{tabular}
\caption{Number of train, development, and test sentences for the out-of-domain experiments}\label{tab:out_of_domain}
\end{table}

\subsection{Dictionaries}
For in-domain translation, we use the ground-truth bilingual dictionaries provided in \citet{conneau2017word}.\footnote{\url{https://github.com/facebookresearch/MUSE##ground-truth-bilingual-dictionaries}} These dictionaries were built using Meta's  internal translation tool and were designed to handle polysemy of words.

For out-of-domain translation we build dictionaries based on the training data, using the method suggested by \citet{shi2021bilingual}. The process is explained in detail in Section~\ref{sec:out}.

\subsection{Prompting Formulation}

Given an instance to translate, we prepend 4 demonstrations to it. Each demonstration consists of 3 parts: (a) the source sentence (along with the translation instruction and the target language): ``\textit{Translate the following sentence to English: <source-sentence>}''; (b) the dictionary-based word-level translations: ``\textit{In this context, the word X means A; the word Y means B,C,D.}''; (c) the translation to the target language: ``\textit{The full translation to English is: <target-sentence>.}''
For the current instance to be translated, part (c) does not include the translation itself, and the model is expected to generate the translation into the target language. See a full example in Figure~\ref{fig:ex}.

To extract the dictionary hints, we look up each of the source words in the dictionary and if an exact match is found,\footnote{We also consider matching after lemmatization (using Stanza~\cite{qi2020stanza}), but since our preliminary experiments find no meaningful difference between these methods, we choose exact match.} we provide the respective dictionary translation(s) as the hint(s). We do not provide any hints for the 500 most frequent source words (based on the development set) as we assume that those are easier for the model to learn and might incorporate noise into the model. In some rare cases where we have more than 3 possible translations for a source word, we choose 3 of them randomly. 

\paragraph{Baselines}

For baselines, we use the same prompt format but without providing the dictionary-based word-level translations. Here, each demonstration consists of only two parts: (a) the source sentence (along with the translation instruction and the target language); and (b) the translation to the target language. See Figure~\ref{fig:ex} for an example -- for the baseline we omit the blue parts (dictionary-based translations).

To ensure a fair comparison, we select the demonstrations randomly from the development set, and consistently use the same demonstrations for all test sentences and across all models.

    
    

\section{Improving Translation Performance}
\label{sec:improving}

In this section, we show the effect of \method{} in the two challenging translation settings: low-resource MT and out-of-domain MT. In both of them we expect to see significant improvements as external information has the potential to fill the gap of missing relevant pretraining data.

\subsection{Low-resource MT}


In Table~\ref{tab:indomain_results}, we report the results for low-resource languages of \method{} vs. the baseline. As described in Section~\ref{sec:langs}, we select 10 languages on which OPT performs moderately well as a proxy for low-resource languages.  

We experiment with OPT and BLOOM for both translation directions (from and into English) across the 10 languages. On average, we gain an improvement of 0.9 BLEU points with OPT and 1.1 BLEU points with BLOOM. Interestingly, OPT performance improves more when translating from English, while Bloom performance improves more when translating into English.

\begin{table*}[h!]
\centering
\scalebox{0.8}{
\begin{tabular}{l|llr|llr}
 \toprule
 \textbf{Language}&\textbf{Baseline}&\textbf{\method{}} & \textbf{Delta}&\textbf{Baseline}&\textbf{\method{}} & \textbf{Delta}\\ 
 &&\textbf{on OPT}&&&\textbf{on Bloom}& \\
\midrule
    \rowcolor{Gray}
    \textbf{Cat-Eng} & \textbf{37.80} & 37.61 & -0.19 & \textbf{46.16} & 45.88 & -0.28\\
     \textbf{Eng-Cat} & 17.86 & \textbf{19.08} & 1.22 & \textbf{40.79} & \textbf{40.79} & 0.00\\
      \rowcolor{Gray}
     \textbf{Hrv-Eng} & 30.72 & \textbf{31.59} & 0.87 & 23.72 & \textbf{25.11}& 1.39\\
     \textbf{Eng-Hrv} & 11.55 & \textbf{12.56} & 1.01 &7.94 & \textbf{9.21}& 1.27\\
       \rowcolor{Gray}
      \textbf{Dan-Eng} & 42.39 & \textbf{42.82} & 0.43 & 34.30 & \textbf{36.87}& 2.57\\
     \textbf{Eng-Dan} & 26.63 & \textbf{27.04} & 0.41& 16.55 & \textbf{19.58}& 3.03\\
      \rowcolor{Gray}
     \textbf{Nld-Eng} & 27.18 & \textbf{27.49} & 0.31& 25.16 & \textbf{25.99}& 0.83\\
     \textbf{Eng-Nld} &  15.98 & \textbf{16.66} & 0.68 & 12.49 & \textbf{13.40}& 0.91\\
     \rowcolor{Gray}
     \textbf{Tgl-Eng} & 31.21 & \textbf{32.12} & 0.91 & 16.95 & \textbf{20.17}& 3.22\\
     \textbf{Eng-Tgl} & 14.05 & \textbf{14.78} & 0.73 & 8.13 & \textbf{9.57}& 1.44\\

     \rowcolor{Gray}
      \textbf{Ind-Eng} & 31.30 & \textbf{32.20} & 0.90 & \textbf{42.55} & 42.18& -0.37\\
     \textbf{Eng-Ind} & 16.03 & \textbf{18.18} & 2.15& \textbf{42.26} & 41.98& -0.28\\

     \rowcolor{Gray}
      \textbf{Ita-Eng} & \textbf{30.03} & 29.84 & -0.19 & 30.60& \textbf{30.95}& 0.35\\
     \textbf{Eng-Ita} & 18.99 & \textbf{19.49} & 0.50 & 19.40& \textbf{19.85}& 0.45\\

      \rowcolor{Gray}
      \textbf{Msa-Eng} & 27.08 & \textbf{28.80} & 1.72 & 42.31 & \textbf{42.47}& 0.16\\
     \textbf{Eng-Msa} & 10.95 & \textbf{12.98} & 2.03 & 30.92 & \textbf{31.30}& 0.38\\

     \rowcolor{Gray}
      \textbf{Nob-Eng} &  38.45& \textbf{39.85} & 1.40 &  30.68 & \textbf{32.73}& 2.05\\
     \textbf{Eng-Nob} & 20.99 & \textbf{22.34} & 1.35 &  13.86 & \textbf{15.38}& 1.52\\

     \rowcolor{Gray}
      \textbf{Slk-Eng} & 24.53 & \textbf{27.12} & 2.59 &  20.15& \textbf{22.33}& 2.18\\
     \textbf{Eng-Slk} & \textbf{6.28} & 5.71 & -0.57& 6.87 & \textbf{8.21}& 1.34\\

  \bottomrule
 \end{tabular}}
\caption{Comparing baseline and \method{} translation results for different low-resource to English and English to low-resource language pairs.}\label{tab:indomain_results}
\end{table*}


\subsection{Out-of-domain MT}
\label{sec:out}

We also study how \method{} performs in out-of-domain prompting-based translation.  This setting is especially interesting since incorporating external information has the potential to result in significant gains when dealing with out-of-domain data.

In these experiments, we translate medical, law, Koran, and IT texts. Despite the possibility that LLMs are trained also on similar domains, we get that the translation quality in these domains is still lacking, as can be seen in the baseline results in Table~\ref{tab:out} (second raw). This is probably the case since LLMs are less likely to have observed sufficient monolingual data for these specialized domains, in contrast to Wikipedia-style data, for example. Sentences from these domains might require translating rare technical terms and idiosyncrasies which present unique challenges even for supervised neural MT models that are well-trained and suited explicitly for translation ~\cite{koehn2017six}.

In this setting, we do not assume that a comprehensive and accurate dictionary specific to a particular domain is available. Such dictionaries can be difficult to obtain, and in some cases may not even exist. Instead, we assume that there is some parallel data available for each domain, and we use this data to create a domain-specific dictionary.

To extract the domain-specific dictionary, we use a combination of word alignment and a fully unsupervised bilingual lexicon induction method. Specifically, we first run the SimAlign algorithm~\cite{sabet2020simalign} on the parallel data, and then apply the method proposed by~\citet{shi2021bilingual}. \citet{shi2021bilingual} propose a to estimate $p(s,t)$ via a smoothed matched ratio between source word $s$ and target word $t$ in their parallel data and align a source word $s$ to the target word
$t$ with the highest $p(s,t)$. For more information, please refer to the paper.
Since this method does not take into account the fact that words can have multiple meanings and translates each source word to only one target word, we modify the algorithm to consider all target words $t$ that have a probability $p(s,t) \ge \lambda$ for each source word $s$.\footnote{Based on our result on development set, we choose $\lambda=0.1$.} 

We compare \method{} with the baseline model (based on Bloom\footnote{We choose Bloom for this experiment as our initial experiments indicate that it is more effective for De-En compared to OPT.}), and with two other models: RePP \cite{sun2022zero} and kNN-MT \cite{khandelwal2020nearest}. RePP is a supervised machine translation system that utilizes bilingual phrase-level translation at test time, improving translation quality during inference through the use of retrieved phrase-level cues. kNN-MT is also based on a supervised MT model and uses the in-domain data for retrieval during inference. Then, the MT output is interpolated with the distributions of the retrieved tokens. 

The results are listed in Table~\ref{tab:out}. The improvement of \method{} over the baseline is striking -- we get that \method{} outperforms the baseline by $9.4$ BLEU points on average. Additionally, \method{} also outperforms RePP by a large margin -- $3.3$ BLEU points on average. As for kNN-MT -- \method{} outperforms it for the KORAN domain. The superiority of kNN-MT over \method{} for the other domains is likely due to a combination of two key-properties: (a) kNN-MT is based on retrieval from a corpus. This allow\pdfoutput
=1s for full style change when needed, which is useful for these domains, as some of them are very patten based (e.g. medical). However, our model is focused on more limited word-level alternations; (b) kNN-MT is much more expensive to run than \method{}. 


\begin{table}[h!]
\centering
\resizebox{\columnwidth}{!}{
\begin{tabular}{l|llll|l}
 \toprule
\bf{Method}& \bf{MED.} & \bf{LAW} & \bf{IT}   &  \bf{KORAN}  &  \bf{Avg.}  \\ 
  \midrule
    Baseline & 37.33 &  35.30 &  23.14 &  16.89 & 28.16\\
  \midrule
  Repp & 44.18& 45.87& 32.44& 14.32& 34.20\\
    \midrule

  \method{} & 50.38 &  45.92 &  33.58 &  20.34 &  37.56  \\
 
    \midrule
     \hline
  \knnmt & 54.54 & 61.11 & 48.63 & 19.22 & 45.87\\
  \bottomrule
  
 \end{tabular}
 }
\caption{Out-of-domain experiment results for medical, law, IT, and Koran domains on $De-En$  data from \citet{aharoni-goldberg-2020-unsupervised}. The baseline and the \method{} are based on the Bloom model.}\label{tab:out} 
\end{table}






\section{Analysis}
\label{sec:analysis}

In this section we present a detailed analysis of \method{}. We provide different hints in several different settings, as explained below, and analyze the resulting outputs.

Naturally, not all source word types have a match in the dictionary. We study the effect of word type coverage in section \ref{sec:coverage}. Additionally, not all respective translations from the dictionary appear in the target reference. In Section \ref{sec:gold} we present an oracle experiment where we only provide gold hints, i.e., hints that are included in the reference translation, to get a sense of the upper bound of \method{}. Finally, to get a better understanding of the generated output and model behavior, we look at a selection of input/output examples in Section~\ref{sec:ex}.

\subsection{Type Coverage Exploration}\label{sec:coverage}
Not all source words have a corresponding entry in the dictionary. The statistics for word token and word type coverage for each language pair in our dictionary are presented in Table~\ref{tab:coverage}. We report both tokens and types in order to get the full picture of word coverage -- word type coverage is the most representative, since once we have a specific word in the dictionary, it will be covered for all instances. Word token coverage is relevant in order to estimate the percent of word instances that are covered in practice. 

\begin{figure}[!hbt]
\centering
\begin{subfigure}{0.40\textwidth}
  \centering
  \includegraphics[width=\linewidth]{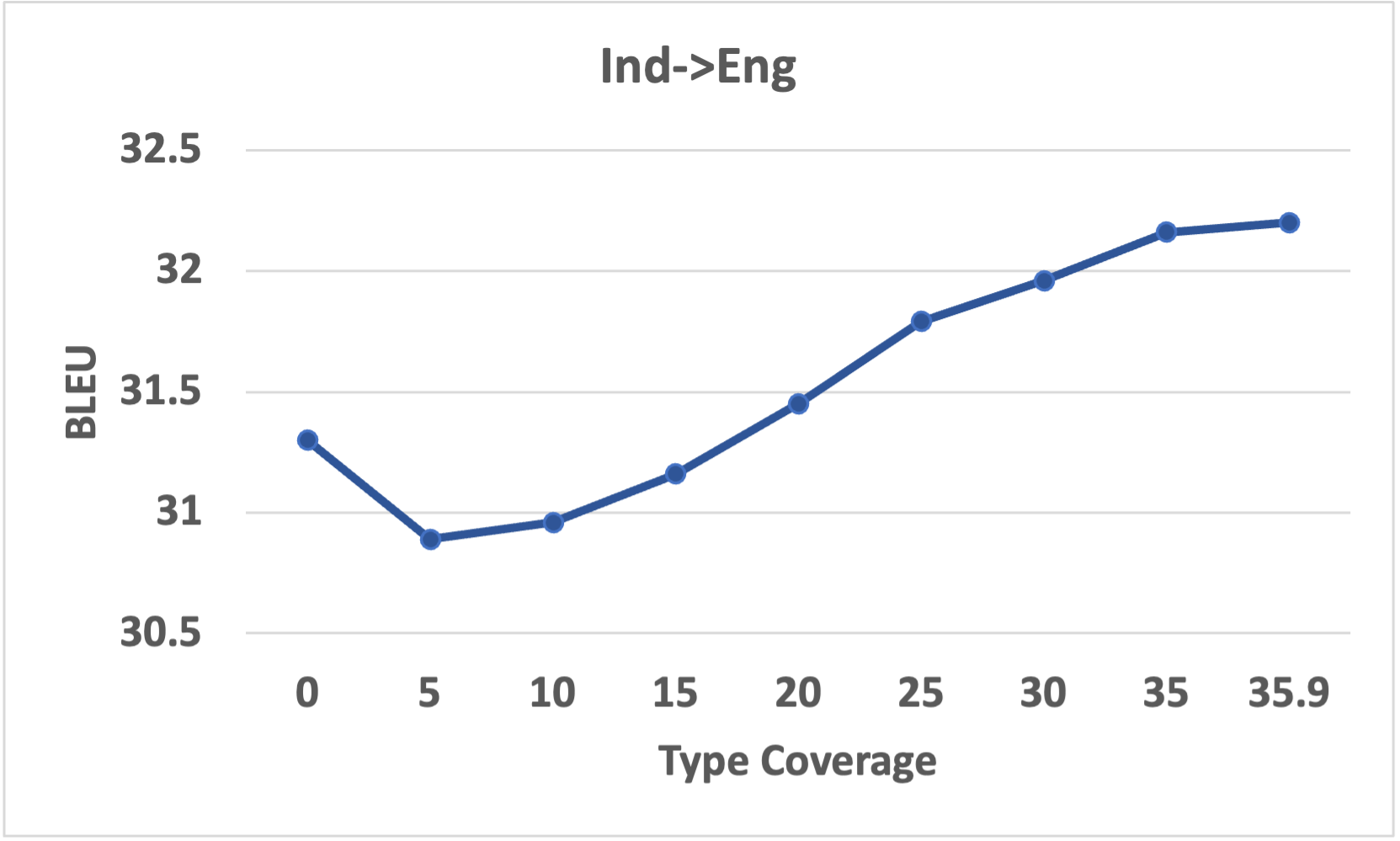}
\end{subfigure}%
\vfil
\begin{subfigure}{0.40\textwidth}
  \includegraphics[width=\linewidth]{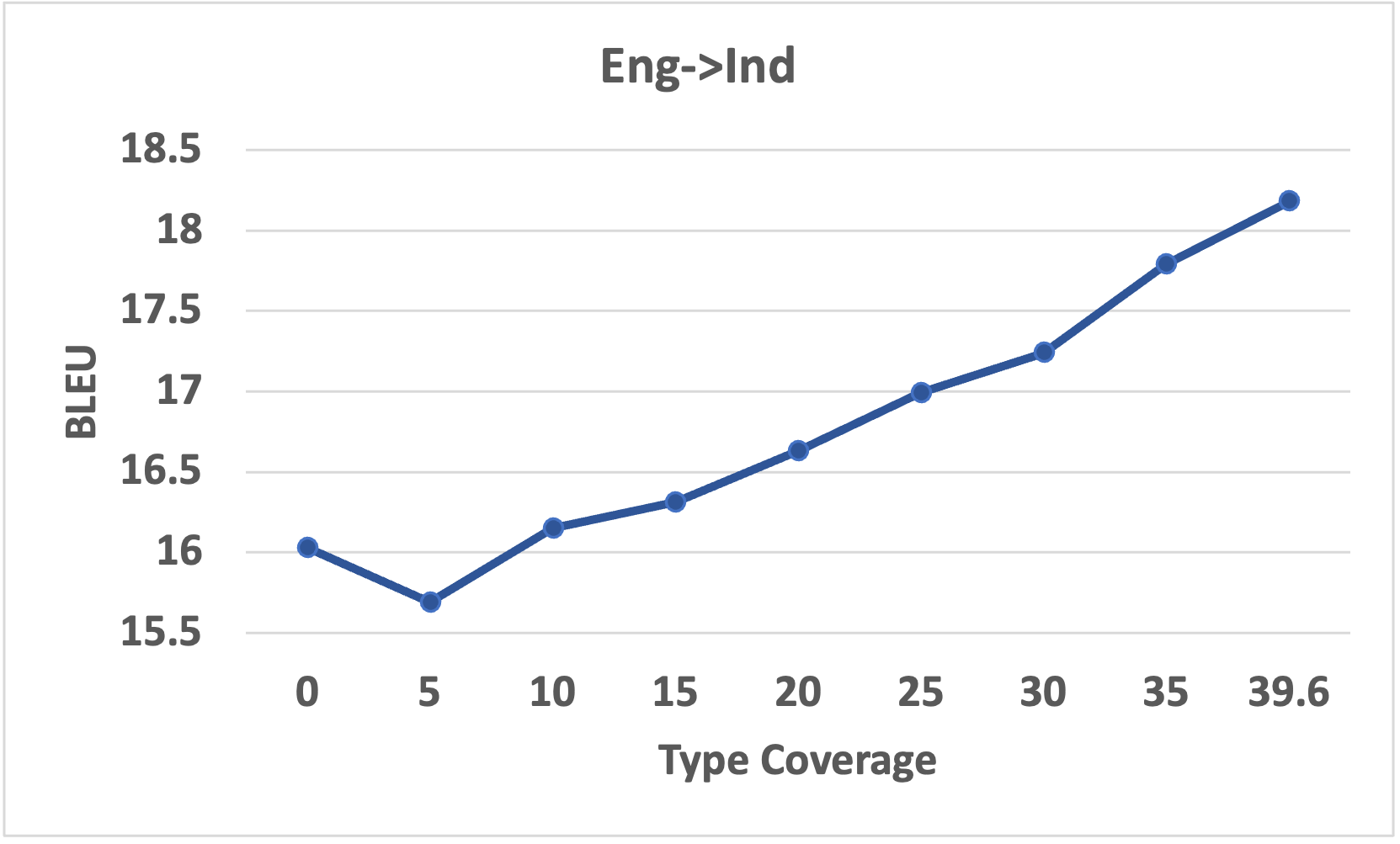}
\end{subfigure}
\begin{subfigure}{0.40\textwidth}
  \includegraphics[width=\linewidth]{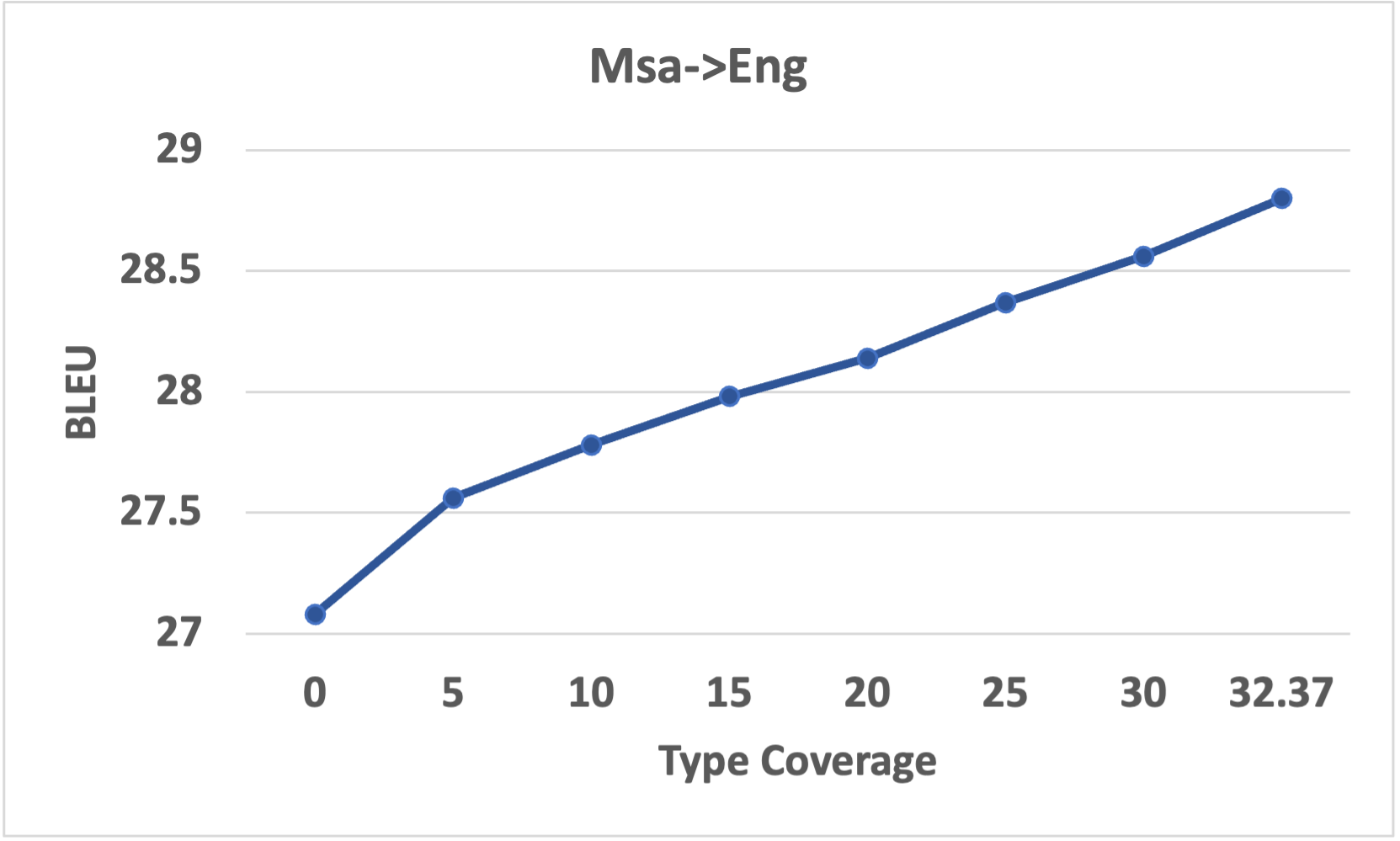}
\end{subfigure}
\begin{subfigure}{0.40\textwidth}
  \includegraphics[width=\linewidth]{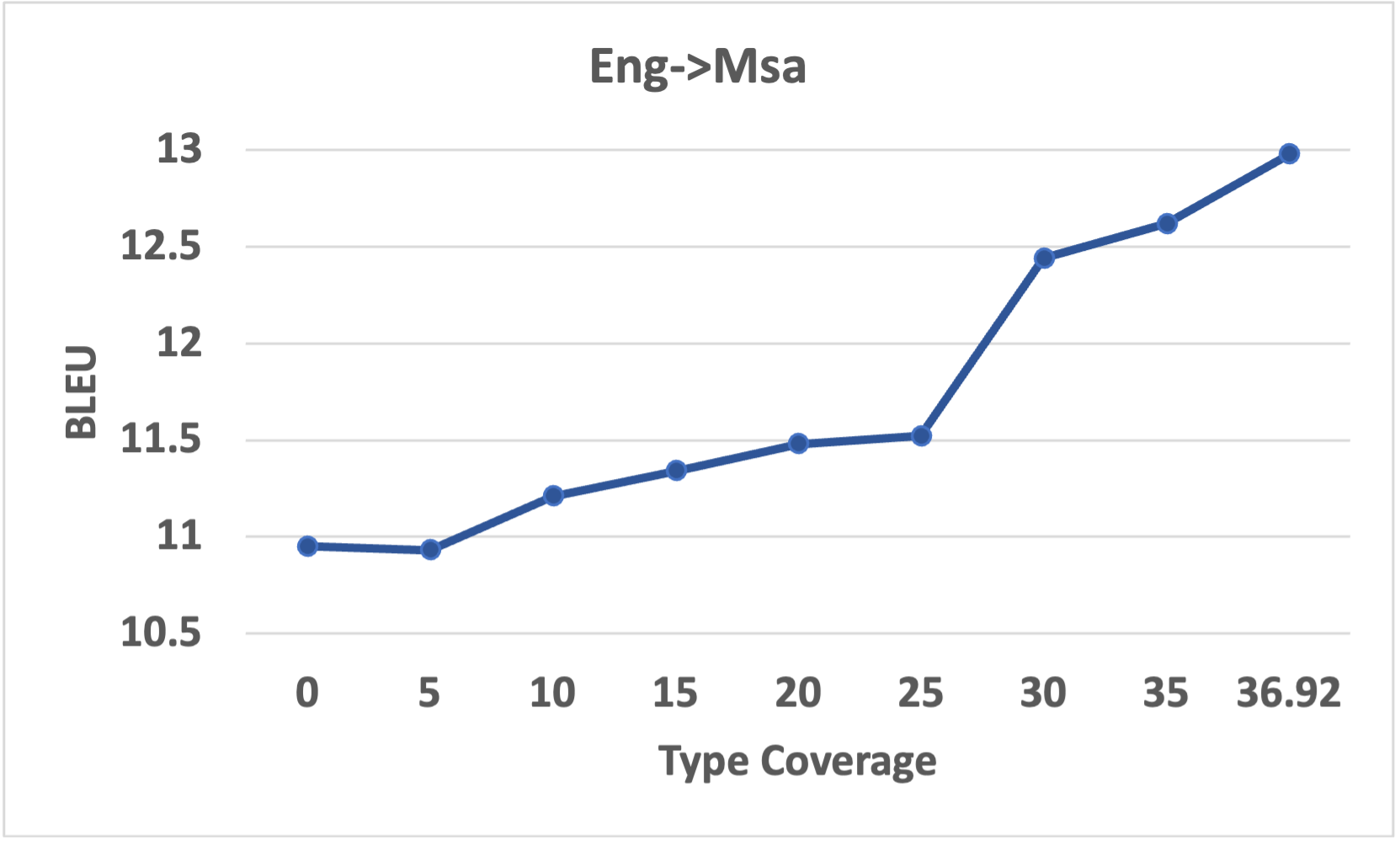}
\end{subfigure}
\caption{Effect of Source Type Coverage on BLEU Improvement Using \method{} based on OPT for English-Indonesian and English-Malay Language Pairs. The baseline results are obtained at $0\%$ type coverage.  }  \label{fig:type_coverage}
\end{figure}

To examine the relationship between BLEU improvement and source word type coverage, we start with the original dictionary and gradually remove random entries to obtain a dictionary version with lower coverage rates. We use a range of coverage rates, from 0\% (no dictionary used) to the full coverage rate (typically around 35\%), with steps of 5\%. We then run our method with these modified versions of the dictionary and report the results. We perform this experiment with two language pairs (Eng$\leftrightarrow$Ind and Eng$\leftrightarrow$Msa) on OPT and present the results in Figure~\ref{fig:type_coverage}. The model performs better than the baseline when the word type coverage is above a certain threshold ($20\%$ for Ind$\rightarrow$Eng and $5-10\%$ for other pairs). Additionally, performance consistently improves as the coverage rate increases. This shows that higher dictionary coverage leads to better results, supporting the utility of \method{} and suggesting that improved dictionary learning is an interesting direction for future work.

\begin{table}[h]
\centering
\scalebox{0.8}{
\begin{tabular}{l|l|l}
 \toprule

 \textbf{Language}& \textbf{Token}  & \textbf{Type}   \\
   \textbf{Pair}& \textbf{Coverage}& \textbf{Coverage} \\
\midrule
    \rowcolor{Gray}
    \textbf{Cat-Eng} & 37.17 & 31.98\\
     \textbf{Eng-Cat} & 40.37 & 34.76\\
      \rowcolor{Gray}
     \textbf{Hrv-Eng} & 25.23& 22.59 \\
     \textbf{Eng-Hrv} & 43.99& 37.52\\
       \rowcolor{Gray}
      \textbf{Dan-Eng} & 38.97& 34.89\\
     \textbf{Eng-Dan} & 47.74 & 41.48\\
      \rowcolor{Gray}
     \textbf{Nld-Eng} & 43.27 & 37.67\\
     \textbf{Eng-Nld} & 47.97 & 41.80\\
     \rowcolor{Gray}
     \textbf{Tgl-Eng} & 31.49 & 25.13\\
     \textbf{Eng-Tgl} &41.32 & 34.58\\

     \rowcolor{Gray}
      \textbf{Ind-Eng} & 44.31 & 35.98\\
     \textbf{Eng-Ind} & 46.11 & 39.60\\

     \rowcolor{Gray}
      \textbf{Ita-Eng} & 43.40 & 37.90\\
     \textbf{Eng-Ita} & 46.61 & 40.31\\

      \rowcolor{Gray}
      \textbf{Msa-Eng} & 39.94 & 32.37\\
     \textbf{Eng-Msa} & 43.41 & 36.92\\

     \rowcolor{Gray}
      \textbf{Nob-Eng} &  38.42 & 33.63\\
     \textbf{Eng-Nob} & 46.40 & 39.75\\

     \rowcolor{Gray}
      \textbf{Slk-Eng} & 30.32 & 26.94\\
     \textbf{Eng-Slk} &48.03 & 41.20 \\

  \bottomrule
 \end{tabular}}
\caption{Dictionary Coverage Rates for Tokens and Types in different Language Pairs.}\label{tab:coverage}
\end{table}

\subsection{Gold Hints}\label{sec:gold}



To get a better understanding of the limits of \method{}, we experiment with using only gold hints from the dictionary, i.e., hints that are included in the reference translation. This experiment provides us with the expected upper bound performance of incorporating dictionary entries into the prompts.

For each source word with a dictionary match, we provide the model only with a single translation -- the one that appears in the reference target sentence, if at all.\footnote{In rare cases where several dictionary translations to the same token are included in the reference, we do not provide a hint for that word.}
This is in contrast to the usual use of \method{} where we provide all the available dictionary translations.




Table~\ref{tab:gold} presents the results. It shows that if \method{} has access to the oracle translations, it can improve the BLEU score by another a 1.1 points on average for OPT.

\begin{table}[h]
\centering
\scalebox{0.8}{
\begin{tabular}{l|lll}
 \toprule

 \textbf{Language}& \textbf{Baseline}  & \textbf{\method{}}  & \textbf{\method{} +}  \\
   \textbf{Pair}& & & \textbf{Gold hints} \\
\midrule
    \rowcolor{Gray}
    \textbf{Cat-Eng} & 37.80 & 37.61 & 38.62\\
     \textbf{Eng-Cat} & 17.86 & 19.08 & 20.97\\
      \rowcolor{Gray}
     \textbf{Hrv-Eng} & 30.72 & 31.59 & 32.93 \\
     \textbf{Eng-Hrv} & 11.55 & 12.56 & 13.21\\
       \rowcolor{Gray}
      \textbf{Dan-Eng} & 42.39 & 42.82 & 43.59\\
     \textbf{Eng-Dan} & 26.63 & 27.04 & 28.66\\
      \rowcolor{Gray}
     \textbf{Nld-Eng} & 27.18 & 27.49 & 28.20\\
     \textbf{Eng-Nld} &  15.98 & 16.66 & 17.13\\
     \rowcolor{Gray}
     \textbf{Tgl-Eng} & 31.21 & 32.12 & 33.32\\
     \textbf{Eng-Tgl} & 14.05 & 14.78 & 16.11\\

     \rowcolor{Gray}
      \textbf{Ind-Eng} & 31.30 & 32.20 & 33.20\\
     \textbf{Eng-Ind} & 16.03 & 18.18 & 18.93\\

     \rowcolor{Gray}
      \textbf{Ita-Eng} & 30.03 & 29.84 &  30.27\\
     \textbf{Eng-Ita} & 18.99 & 19.49 &  20.81\\

      \rowcolor{Gray}
      \textbf{Msa-Eng} & 27.08 & 28.80 & 29.63\\
     \textbf{Eng-Msa} & 10.95 & 12.98 & 13.87\\

     \rowcolor{Gray}
      \textbf{Nob-Eng} &  38.45& 39.85 & 40.84\\
     \textbf{Eng-Nob} & 20.99 & 22.34 & 23.87\\

     \rowcolor{Gray}
      \textbf{Slk-Eng} & 24.53 & 27.12 &  28.78\\
     \textbf{Eng-Slk} & 6.28 & 5.71 & 7.48\\

  \bottomrule
 \end{tabular}}
\caption{Translation results of \method{} when we provide all the available dictionary translations vs only gold hints (based on OPT). The Second and Third columns are the same as Table~\ref{tab:indomain_results}.}\label{tab:gold}
\end{table}

\subsection{Output Analysis}
\label{sec:ex}

In this section we select a few representative input/output examples to demonstrate the strengths and weaknesses of \method{}.

Figure~\ref{fig:good_hints} exemplifies the way \method{} helps the model generate the correct translation. Such examples are common and are the source of the BLEU improvement we see in Section~\ref{sec:improving}. In this example the hints in boldface for the three words "serangga" (insects), "melipat" (fold), "capung" (dragonflies) have helped the system generate a better translation.

\begin{figure}[!hbt]
    \centering
    \begin{tcolorbox}[boxsep=0pt,left=0pt,colframe=white]
    \begin{tabular}{p{\linewidth}}
    
    \textbf{Source:} Saat ini, satu-satunya serangga yang tidak dapat melipat sayapnya adalah capung dan lalat capung. \\ 
    \textbf{Reference:} Today, the only insects that cannot fold back their wings are dragon flies and mayflies.\\ 
    \textbf{Baseline:} At present, the only bird that cannot fly is the capuchin monkey. \\ 
    \textbf{\hint{}:}  In this context, the word "serangga" means \textbf{"insects"}, "insect", "bugs"; the word "melipat" means \textbf{"fold"}, "folding"; the word "capung" means "dragonfly", \textbf{"dragonflies"}; the word "lalat" means "fly", "flies". \\

    \textbf{\method{}:} At the moment, the only insects that cannot fold their wings are dragonflies and damselflies.  \\ \\ 

    \end{tabular}
    \end{tcolorbox}    
    \caption{An example where hints are used by the model to generate a better translation.}  
    \label{fig:good_hints}
\end{figure}

The next examples show failure cases of \method{}: the model's output initially matches the reference, but when provided with the hints, some tokens are mistranslated either because the desired translation is not included in the hints (Figure~\ref{fig:bad_hints_style}) or it is included but the model picks some other provided hint (Figure~\ref{fig:bad_hints_option}). In both of these examples the generated output is still a correct translation but there is a style mismatch between the new output and the reference. 

\begin{figure}[!hbt]
    \centering
    \begin{tcolorbox}[boxsep=0pt,left=0pt,colframe=white]
    \begin{tabular}{p{\linewidth}}
    
    \textbf{Source:} Pelancong mungkin akan menghadapi binatang hama yang tidak mereka kenal di kampung halaman mereka. \\ 
    \textbf{Reference:} Travellers may encounter animal pests that they are not familiar with in their home regions. \\ 
    \textbf{Baseline:} Tourists may encounter wild animals that they are not familiar with in their home villages. \\ 
    \textbf{\hint{}:} In this context, the word "menghadapi" means \textbf{"facing"}; the word "binatang" means "beast", "beasts", "animals", "animal"; the word "kenal" means \textbf{"know"}; the word "kampung" means "kampong", "kampung", "village".\\

    \textbf{\method{}:}  Tourists may face wild animals that they do not know in their village.  \\ \\



    \end{tabular}
    \end{tcolorbox}    
    \caption{An example where hints cause a mistranslation because the desired translation is not included in the hints.} 
    \label{fig:bad_hints_style}
\end{figure}

\begin{figure}[!hbt]
    \centering
    \begin{tcolorbox}[boxsep=0pt,left=0pt,colframe=white]
    \begin{tabular}{p{\linewidth}}
    
    \textbf{Source:}  Semua binatang yang awalnya tiba di kepulauan ini datang dengan cara berenang, terbang, atau mengambang.\\ 
    \textbf{Reference:} All animals that originally arrived in the islands came here either by swimming, flying or floating. \\ 
    \textbf{Baseline:}   All animals that originally arrived in the islands came here either by swimming, flying or floating. \\ 
    \textbf{\hint{}:} In this context, the word "binatang" means "beast", "beasts", "animals", "animal"; the word "awalnya" means "initially", "originally"; the word "kepulauan" means \textbf{"archipelago", "islands"}; the word "datang" means "comming", "come", "coming". \\

    \textbf{\method{}:}  All the animals that first arrived in this archipelago came by beaching, flying, or swimming. \\ \\ 

    \end{tabular}
    \end{tcolorbox}    
    \caption{An example where hints cause a mistranslation because multiple hints are provided to the model and it does not pick the desired one.  }  
    \label{fig:bad_hints_option}
\end{figure}

The model often makes use of the hints to generate the output. The example in Figure~\ref{fig:ignore_hints} shows a relatively rare case where the baseline output is incorrect and the hints are not used in the new output even though they match the translation in the reference. In this example the model chooses to ignore the hints even though they could help in improving the translation. 

\begin{figure}[!hbt]
    \centering
    \begin{tcolorbox}[boxsep=0pt,left=0pt,colframe=white]
    \begin{tabular}{p{\linewidth}}

    \textbf{Source:}  Ukuran atom luar biasa kecil sehingga titik di titik ujung kalimat ini dapat memuat beberapa triliun atom. \\ 
    \textbf{Reference:} Atoms are so incredibly tiny that trillions of them could fit into the period at the end of this sentence. \\ 
    \textbf{Baseline:}  The size of the atom is very small so that the tip of the pen can contain several trillion atoms. \\ 
    \textbf{\hint{}:} In this context, the word "atom" means "atoms", "atom"; the word "kalimat" means \textbf{"sentence"}, "sentences"; the word "triliun" means "trillion". \\

    \textbf{\method{}:}   The size of the atom outside the nucleus is small enough to contain several trillion atoms. \\ \\

    \end{tabular}
    \end{tcolorbox}    
    \caption{An example where the model chooses to ignore the hints even though they could help in improving the translation.}  
    \label{fig:ignore_hints}
\end{figure}

\section{Controllability}
\label{sec:control}

We now turn to study the extent to which \method{} controls the final translation. In the previous section we observe that incorporating hints indeed helps the model generate better translations, but now we are interested to explore the actual direct effect the hints have on the translation. 

To evaluate the level of controllability of \method{} we look at the percentage of times the system uses a suggested hint in the generated output. This should, of course, be compared with the respective percentage for the baseline, since the model might use a suggested hint even in baseline setting, when the hint is not provided. In case multiple translation hints are available for a single source token -- if any of the hints appears in the target sentence, we consider it as \textit{hit}. Multiple \textit{hits} for a single source token are counted as one. 

We compare these results with two control experiments: in \textit{random hint}, we choose a single dictionary translations (if available) for each word at random and provide it as the only hint for that word; in \textit{false hint}, we shuffle the target side of the dictionary to obtain a false dictionary, and proceed similarly. 

The results are presented in Table~\ref{tab:control}. As expected, we get that \method{} is able to control the final input, with significant gap over the percentage of the baseline (\textit{dictionary hint}). The highest controllability level is achieved when providing only the gold hint. Surprisingly, for the \textit{false hint} setting, we get almost no change with respect to the baseline, suggesting that our method is resilient to false word-level hints.

\begin{table*}[!hbt]
\centering
\scalebox{0.75}{
\begin{tabular}{l|lll|lll||lll|lll|}
 \toprule
 \textbf{Language}&\multicolumn{3}{c|}{Dictionary Hint} & \multicolumn{3}{c||}{Gold Hint} &  \multicolumn{3}{c|}{Random Hint} & \multicolumn{3}{c|}{False Hint}\\
 
 \textbf{Pair} & \textbf{Baseline}  & \textbf{\method{}}  & \textbf{Delta} & \textbf{Baseline}  & \textbf{\method{}}  & \textbf{Delta} & \textbf{Baseline}  & \textbf{\method{}}  & \textbf{Delta} & \textbf{Baseline}  & \textbf{\method{}}  & \textbf{Delta} \\
\midrule

\rowcolor{Gray}
 \textbf{Cat-Eng} & 64.4 & 75.7 & 11.3 & 75.4 & 94.2 & 18.8 & 61.5 & 73.3 & 11.8 & 0.1 & 0.3 & 0.2 \\
\textbf{Eng-Cat} & 39.5 & 57.8 & 18.3 & 45.5 & 71.9 & 26.4 & 34 & 52 & 18 & 0.2 & 0.4 & 0.2 \\

 \rowcolor{Gray}
\textbf{Hrv-Eng} & 59.1 & 71.7 & 12.6 & 66.6 & 91.3 & 24.7 & 50.1 & 66.8 & 16.7 & 0.2 & 0.5 & 0.3 \\
\textbf{Eng-Hrv} & 27.5 & 41 & 13.5 & 31.5 & 59.4 & 27.9 & 16.6 & 31.2 & 14.6 & 0.2 & 0.5 & 0.3 \\

 \rowcolor{Gray}
\textbf{Dan-Eng} & 71.5 & 81.2 & 9.7 & 79 & 94.2 & 15.2 & 57.1 & 68.6 & 11.5 & 0.1 & 0.2 & 0.1 \\
\textbf{Eng-Dan} & 50.5 & 67.6 & 17.1 & 53.8 & 78.9 & 25.1 & 30.3 & 47.2 & 16.9 & 0.1 & 0.2 & 0.1 \\

 \rowcolor{Gray}
\textbf{Nld-Eng} & 54.3 & 80.7 & 26.4 & 61.5 & 93.4 & 31.9 & 46.7 & 76.4 & 29.7 & 0 & 0.1 & 0.1 \\
\textbf{Eng-Nld} & 48.2 & 59.7 & 11.5 & 51.6 & 76.1 & 24.5 & 33.3 & 49.8 & 16.5 & 0.1 & 0.2 & 0.1 \\

 \rowcolor{Gray}
\textbf{Tgl-Eng} & 49.1 & 62.4 & 13.3 & 60.4 & 86.3 & 25.9 & 39.3 & 58 & 18.7 & 0.2 & 0.9 & 0.7 \\
\textbf{Eng-Tgl} & 25.5 & 41.3 & 15.8 & 31.7 & 60.3 & 28.6 & 18.2 & 57.3 & 39.1 & 0.3 & 0.7 & 0.4 \\

 \rowcolor{Gray}
\textbf{Ind-Eng} & 54.3 & 69.6 & 15.3 & 59.1 & 83.3 & 24.2 & 32 & 51.9 & 19.9 & 0.3 & 0.3 & 0 \\
\textbf{Eng-Ind} & 30.8 & 51 & 20.2 & 35.2 & 66.8 & 31.6 & 20.1 & 40.6 & 20.5 & 0.3 & 0.5 & 0.2 \\

 \rowcolor{Gray}
\textbf{Ita-Eng} & 68.0 & 74.7 & 6.7 & 81.1 & 93.9 & 12.8 & 54.6 & 63.9 & 9.3 & 0.2 & 0.2 & 0 \\
\textbf{Eng-Ita} & 53.5 & 61.1 & 7.6 & 59.2 & 78.5 & 19.3 & 33.3 & 49.8 & 16.5 & 0.1 & 0.2 & 0.1 \\

 \rowcolor{Gray}
\textbf{Msa-Eng} & 43.9 & 65.2 & 21.3 & 50.8 & 82.7 & 31.9 & 30.1 & 53.1 & 23 & 0.2 & 0.9 & 0.7 \\
\textbf{Eng-Msa} & 21.1 & 41 & 19.9 & 26.3 & 59.2 & 32.9 & 17.8 & 44.3 & 26.5 & 0.2 & 0.9 & 0.7 \\

 \rowcolor{Gray}
\textbf{Nob-Eng} & 68.6 & 78.8 & 10.2 & 79.8 & 93.9 & 14.1 & 56.2 & 68.2 & 12 & 0.1 & 0.1 & 0 \\
\textbf{Eng-Nob} & 47.5 & 64.8 & 17.3 & 53.1 & 79.1 & 26 & 32 & 48.9 & 16.9 & 0.1 & 0.3 & 0.2 \\

 \rowcolor{Gray}
\textbf{Slk-Eng} & 49.7 & 69.2 & 19.5 & 59.9 & 87.5 & 27.6 & 38.4 & 63.5 & 25.1 & 0.3 & 0.9 & 0.6 \\
\textbf{Eng-Slk} & 14.3 & 35.3 & 21 & 14.5 & 45.3 & 30.8 & 6.2 & 29.4 & 23.2 & 0.2 & 0.8 & 0.6 \\

  \bottomrule
 \end{tabular}}
\caption{Controllability results: the percentage of times a hint is used in the output for four different hinting strategies. The Delta column is what we are interested in since the model might use a suggested hint even in the baseline setting. Results are based on the OPT model.}\label{tab:control}
\end{table*}

\section{Related Work}
\paragraph{Prompting Language Models for MT} There have been relatively few studies on prompting language models for machine translation. Most research in this area has focused on testing the machine translation ability of large language models using simple prompts like $\{$Source text$\} =\{$Target text$\}$ or  Translate to $\{$language\_name$\}: \{$text$\}$ \cite{brown2020language, lin2021few, zhang2022opt,scao2022bloom, garcia2022using}. \citet{reynolds2021prompt} experiment with different prompt templates, and \citet{garcia2022using}  explore the use of prompts for controlling various aspects of the formality or specific dialect of the output.  There is also a line of work (\citealt{agrawal2022context, vilar2022prompting}) that concentrates on choosing good in-context examples for machine translation. This is in parallel with our work, which utilizes dictionary for better translation.

\paragraph{Using Dictionary in  MT}
Several researchers have investigated the use of dictionaries in supervised machine translation. \citet{zhang2016bridging} propose a method that combines NMT with a bilingual dictionary containing rare or unseen words in the bilingual training data.   \citet{arthur2016incorporating}  present a method for improving the translation of low-frequency  words in NMT by augmenting the system with discrete translation lexicons and using the attention vector to select relevant lexical probabilities. \citet{zhong2020look}  propose a similar method for using  bilingual and monolingual dictionaries to improve NMT. In related research, \citet{hamalainen2019template} use a dictionary to build synthetic parallel data for training a stronger NMT system.

Our work is also related to lexical constraints in MT, which can be classified as hard constraints \cite{hokamp2017lexically, post2018fast}   or soft constraints \cite{song2019code,dinu2019training,chen2021lexical}.

\paragraph{Domain Adaptation for MT} 

Previous efforts have been made to enhance the performance of pre-trained NMT models using out-of-domain bilingual or monolingual datasets. Our method, similar to prevoius works \cite{khandelwal2020nearest, zheng2021non, agrawal2022context}, adapts to the new domain during inference time and does not require additional training. The work most similar to ours is \citet{sun2022zero}, which trains a system that can use bilingual phrase-level translation at test time and improves translation quality during inference by constructing a bilingual phrase-level database and using retrieved phrase-level cues.
 

\section{Conclusion}

We propose a method for incorporating bilingual dictionary information into prompting-based MT by explicitly adding possible word-level translations into the prompt. 

We show that our method, \method{}, improves translation quality for low-resource languages. Additionally, we tackle out-of-domain translation by creating dictionaries in an unsupervised manner and using them with our method to get impressive translation improvements.

Our analyses of \method{} demonstrate the benefits and limitations of the method and quantify the level of controllability the method has on the final translation under different conditions.



\bibliography{anthology,custom}
\end{document}